\def\year{2021}\relax
\documentclass[letterpaper]{article} 
\usepackage{aaai21}  
\usepackage{times}  
\usepackage{helvet} 
\usepackage{courier}  
\usepackage[hyphens]{url}  
\usepackage{graphicx} 
\usepackage{bm}
\usepackage{multirow}
\usepackage{mathtools,amsmath}
\DeclareMathOperator*{\argmax}{argmax}
\usepackage{natbib}  
\usepackage{caption} 
\title{Neural Process for Black-Box Model Optimization Under Bayesian Framework}
\author{
    Zhongkai Shangguan,\textsuperscript{\rm 1} 
    Lei Lin,\textsuperscript{\rm 2}
    Wencheng Wu,\textsuperscript{\rm 2}
    Beilei Xu\textsuperscript{\rm 2}
    \\
}
\affiliations{
    \textsuperscript{\rm 1 \rm 2}Rochester Data Science Consortium, University of Rochester, Rochester, NY 14604 USA\\

    \textsuperscript{\rm 1}zshangg2@ur.rochester.edu
    \textsuperscript{\rm 2}\{Lei.Lin, Wencheng.Wu, Beilei.Xu\}@rochester.edu

}

\begin{document}

\maketitle

\begin{abstract}
There are a large number of optimization problems in physical models where the relationships between model parameters and outputs are unknown or hard to track. These models are named as “black-box models" in general because they can only be viewed in terms of inputs and outputs, without knowledge of the internal workings. Optimizing the black-box model parameters has become increasingly expensive and time consuming as they have become more complex. Hence, developing effective and efficient black-box model optimization algorithms has become an important task. One powerful algorithm to solve such problem is Bayesian optimization, which can effectively estimates the model parameters that lead to the best performance, and Gaussian Process (GP) has been one of the most widely used surrogate model in Bayesian optimization. However, the time complexity of GP scales cubically with respect to the number of observed model outputs, and GP does not scale well with large parameter dimension either. Consequently, it has been challenging for GP to optimize black-box models that need to query many observations and/or have many parameters. To overcome the drawbacks of GP, in this study, we propose a general Bayesian optimization algorithm that employs a Neural Process (NP) as the surrogate model to perform black-box model optimization, namely, Neural Process for Bayesian Optimization (NPBO). In order to validate the benefits of NPBO, we compare NPBO with four benchmark approaches on a power system parameter optimization problem and a series of seven benchmark Bayesian optimization problems. The results show that the proposed NPBO performs better than the other four benchmark approaches on the power system parameter optimization problem and competitively on the seven benchmark problems.
\end{abstract}

\section{Introduction}
A complex physical model can be viewed as a black-box model since it can only be viewed in terms of its inputs (parameters) and outputs (observations) when internal workings are hard to track. Optimizing the parameters of such a black-box model is a common problem in engineering fields \cite{xiao2015optimization, tutorialbbo, zhang2010industrial}. For examples, power system models need regular calibrations to reflect the current power system status, identify and mitigate potential issues \cite{2020HPT-RL}; geometric structure parameter optimization of an antenna aims at reaching an optimal gain in working frequency band \cite{gustafsson2016antenna}. 
In many cases, this optimization process is done manually by experts based on a series of experiments. However, this process requires expert domain knowledge and a number of experiments to be conducted, which can be both expensive and time consuming. In order to reduce human effort and the number of required experiments, automated optimization algorithms with varying computational complexity and scalability have been proposed. 

Conventional automated black-box model optimization algorithms include grid search and random search \cite{bergstra2012random}. In grid search, the problem is defined in a high-dimensional grid space where each grid dimension corresponds to a parameter, and each grid point corresponds to a parameter combination. We then evaluate the model on all parameter combinations defined by the grids, and select the parameter combination that yields the best performance of the model. One drawback of grid search is that the number of grid points grows exponentially as the number and value range of the parameters increase. On the other hand, random search approach can potentially explore the parameter space more extensively through randomly generates parameter combinations in the parameter space. 
However, both of these two methods do not utilize the prior sampled information, so that the search is blind. Recently, more advanced automated optimization algorithms have been introduced, including evolutionary optimization \cite{jade_2018}, population-based optimization \cite{jaderberg2017population}, and Bayesian optimization \cite{frazier2018tutorial}. These algorithms construct the relationship between parameter combinations and the performance of black-box models, and provide guidance for the next selection of parameter combination for evaluation, which make those methods more efficient to find the global optimal parameter combination.

In this paper, we focus on the Bayesian optimization framework. Bayesian optimization implements a surrogate model that predicts the black-box model performance for a specific parameter combination; and an acquisition function, which trades off exploration and exploitation to query the next observation point. An accurate surrogate model that can predict the black-box model performance as well as measure the uncertainty is crucial to the performance of Bayesian optimization.
Gaussian process (GP) has been the most widely used surrogate model for Bayesian optimization \cite{3569} due to its expressiveness, smoothness and well-calibrated uncertainty estimation of the model response.
However, GP has $\mathcal{O}(N^3)$ time complexity, where $N$ is the number of training samples \cite{3569}, so it is computationally expensive. Another drawback of GP is its poor scalability to high parameter dimensions \cite{liu2020gaussian}, i.e., with the increase in the dimension of parameters, the performance of GP becomes worse. Hence, GP cannot be applied to problems that require to query many observations and/or have many parameters. 

To overcome these issues, we propose a new algorithm for black-box model optimization by employing Neural Process as a powerful and scalable surrogate model under the Bayesian optimization framework, namely, Neural Process for Bayesian Optimization (NPBO). Neural Process (NP) \cite{garnelo2018neural} is an algorithm to simulate stochastic process using neural networks (NNs). It combines the advantages of stochastic process and NNs so that it has the ability to capture uncertainty while predicting the black-box model's performance accurately. The performance of the proposed algorithm is evaluated on a power system parameter optimization problem and seven benchmark problems for Bayesian optimization \cite{klein2017robo}. The results show that the proposed NPBO outperforms the other four benchmark approaches including GP-based Bayesian optimization, random forest based Bayesian optimization, Deep Networks for Global Optimization (DNGO) and Bayesian Optimization with Hamiltonian Monte Carlo Artificial Neural Networks (BOHAMIANN) on the power system parameter optimization problem and performs competitively on the benchmark problems.

The rest of the paper is arranged as follows. \textbf{Related Work} introduces state-of-the-art methods in parameter optimization under the Bayesian optimization framework. The methodology of NPBO is introduced in detail in \textbf{Methodology}. We compare and discuss the results of NPBO and the benchmark algorithms in the \textbf{Experiment} section. Finally, the paper concludes our work and discusses future steps.

\section{Related Work}
This section will introduce the framework of Bayesian optimization using GP, then discuss previous proposed approaches \cite{snoek2015scalable, springenberg2016bayesian} that use NNs as alternative surrogate models.

\subsection{Bayesian Optimization Framework}
Bayesian optimization is a state-of-the-art optimization framework for black-box model optimization problems. It has great power in parameter optimization of physical models \cite{duris2020bayesian, building} and hyperparameter optimization in training machine learning (ML) models \cite{chen2018bayesian}. A detailed tutorial can be found in \cite{archetti2019bayesian}.

The aim of Bayesian optimization is to find the input $\bm x$ that maximizes an unknown nonlinear function $f(\bm{x}):R^d \rightarrow R$. It can be formulated as (\ref{equ:bo}):
\begin{equation}\label{equ:bo}
    \hat{\bm x} = \argmax_{\bm x\in A \subset R^d} f(\bm x),
\end{equation}
where $d$ is the dimension of $\bm x$, $A$ is the constrain that defined for $f(\bm x)$, $f(\cdot)$ is the nonlinear function that is expensive to evaluate, and $\hat {\bm x}$ is the estimation of the input parameter. One assumption of Bayesian optimization is that only the outputs $f(\bm x)$ can be observed while its derivatives cannot be obtained. Hence, $f(\cdot)$ is a black-box model and the optimization problem cannot be solved using gradient descent algorithm.
Bayesian optimization repeatedly executes the following steps until a satisfactory input parameter combination is found: (i) fit a surrogate model to the current observations to get a prior distribution; (ii) convert the prior to the posterior distribution and predict where the next input parameter combination is by maximizing an acquisition function; (iii) obtain the observation on the suggested parameter combination and add the result to the observation set. The third step is usually the most expensive step since it needs to generate the observation on the expensive black-box model.

There are two main components in Bayesian optimization: a surrogate model that simulates the black-box model and an acquisition function that trades off exploration and exploitation in order to decide the next query inputs. GP is a classical model that is widely employed as the surrogate model by Bayesian optimization. A GP is defined as a stochastic process indexed by a set $\mathcal{X}\subseteq \mathcal{R}^d:{\{ f(\bm x):\bm x \in \mathcal{X} \}}$ such that any finite number of random variables of the process has a joint Gaussian distribution. Instead of inferring a distribution over the parameters, GP can be used to infer a distribution over the function directly. For example, considering we sample a finite set $D=\{\bm x_1, ..., \bm x_n\}$, $D\in \mathcal{X}$, GP is completely defined by its mean and covariance functions as (\ref{equ:gp}) shows
\begin{equation}
    \label{equ:gp}
    \begin{multlined}
        p(\bm f|D) = \mathcal{N}(\bm f|\bm \mu, \bm K)
    \end{multlined}
\end{equation}
where $\bm f = (f(\bm x_1),...f(\bm x_n))$ is the distribution over the black-box model; $\bm\mu = (m(\bm x_1),..., m(\bm x_n))$ where $m$ is the mean function, and $\bm K=K(\bm x_i, \bm x_j)$ represents the covariance function (also known as kernel) such as Radial basis function kernel \cite{gortler2019visual}. So for a specific $\bm x$, GP predicts a mean and a variance that completely define a Gaussian distribution over $f(\bm x)$, i.e., $f(\bm x) \sim \mathcal{N}(f(\bm x)|m(\bm x), K(\bm x))$. 

Expected improvement (EI) is one of the popular acquisition functions in Bayesian optimization \cite{frazier2018tutorial}. EI is computed as the expectation taken with respect to the posterior distribution. It is defined as:
\begin{equation}\label{equ:ei}
    \begin{multlined}
        \alpha_{EI}(\bm x) := \\ 
        (\mu (\bm x) -\tau)\Phi \left (\frac{\mu (\bm x)-\tau}{\sigma(\bm x)} \right)+\sigma(\bm x) \phi \left (\frac{\mu (\bm x)-\tau}{\sigma(\bm x)}\right)
    \end{multlined}
\end{equation}
where $\Phi$ and $\phi$ represent the cumulative distribution function and probability distribution function of standard normal distribution, $\tau$ is the current best observed outcome, $\mu$ and $\sigma$ are the mean and variance.

\subsection{Bayesian Optimization with Neural Networks}
To address the downsides of the standard GP, i.e., scaling cubically with the number of queried observations and not performing well with high dimensional data, alternative methods based on sparse GP approximations \cite{snelson2006sparse, lazaro2010sparse, mcintire2016sparse} have been proposed. These methods approximate the full GP by using subsets of the original dataset as inducing points to build a covariance function. However, they are not accurate in uncertainty estimation and still have poor scalability in high dimensional parameter space \cite{hebbal2019bayesian}. Since NNs are very flexible and scalable, adapting NNs to the Bayesian optimization framework is highly desirable. However, NNs are deterministic models that do not have the ability to measure the uncertainty. Hence, combining the flexibility and scalability of NNs with well-calibrated uncertainty estimations is crucial in this procedure. 

More recently, Bayesian optimization methods based on NNs have been proposed. \citet{snoek2015scalable} propose a Deep Networks for Global Optimization (DNGO) framework that uses NNs as the feature extractor to pre-process the inputs, and then adapts a Bayesian linear regression (BLR) model to gain the uncertainty. Specifically, the authors first train a deterministic neural network with fully-connected-layers, where the output of the penultimate layer is regarded as many basis functions and the final layer is regarded as only a linear combination of these basic functions. Then they freeze the parameters of neural network and feed the basis functions generated by the penultimate layer to a probabilistic model, i.e., BLR, to measure the uncertainty. This method successfully embeds neural networks into a probability model but uncertainty measurement only takes a small part in the whole procedure, which make it not perform well in uncertainty estimation.

\citet{springenberg2016bayesian} apply a Bayesian Neural Network (BNN) as the surrogate model and train it with stochastic gradient Hamiltonian Monte in the Bayesian optimization framework (BOHAMIANN). In detail, they treat the weights of a neural network as a probability distribution so as to model it in a probabilistic approach. In inference, they sample several models from the probabilistic model to acquire the uncertainty. This method combines the advantages of neural networks and Bayesian models, but in the inference stage, it need to sample several times to obtain the uncertainty. Hence, if only a few models are sampled, the uncertainty estimation will not be accurate; on the other hand, sampling too many models will make the algorithm time consuming.

\section{Methodology}
In this section, we show how NP can be used as the surrogate model of Bayesian optimization. We first summarize the general formalism behind NP, then derive the training process for the proposed NPBO algorithm.

\subsection{Neural Process for Bayesian Optimization}
NP is a neural network based approach to represent a distribution over functions. It builds neural networks to model functions as a random process $f$. Given a set of observations $((x_1, y_1), ..., (x_N, y_N))$, NP first defines $\rho_{x_{1:N}}$ as the marginal distribution of $(f(x_1), ..., f(x_N))$, i.e. 
\begin{equation}\label{equ:4}
    \rho_{x_{1:N}}(y_{1:N}) := \rho_{x_1,...,x_N}(y_1, ...,y_N)
\end{equation}
Assuming that the independence among samples and observation noise exists, i.e. $y_i=f(x_i)+\varepsilon_i \sim \mathcal{N}(f(x_i), \sigma^2(x_i))$, in order to build a model on (\ref{equ:4}), the conditional distribution can be written as 
\begin{equation}\label{equ:5}
    p(y_{1:N}|x_{1:N}, f) = \prod_{i=1}^{N} \mathcal{N}(y_i|f(x_i), \sigma^2(x_i))
\end{equation}
where $p$ denotes the probability distribution. 
Then, in order to build the stochastic process $f$ using neural network, we assume $f(x) = g(x, \bm z)$, where $\bm z$ is a latent vector that is sufficient to represent $f$, and $g$ is a general function defined for $f$. We assume $p(\bm z)$ obeys a multivariate standard normal, then the random process $f$ becomes sampling of $\bm z$. For example, assuming $f$ is a GP, then $\bm z$ should be a vector containing the mean and variance that can fully define the GP. Hence, a NN can be employed to output $\bm z$ so as to model the random process $f$.
Replacing  $f(x)$ with $g(x, \bm z)$, we have 
\begin{equation}\label{equ:6}
    p(\bm z, y_{1:N}|x_{1:N}) = p(z)\prod_{i=1}^{N}\mathcal{N}(y_i|g(x_i, \bm z), \sigma^2(x_i))
\end{equation}

\subsection{Modeling and Training}
\begin{figure*}
    \includegraphics[width=\textwidth, height=5cm]{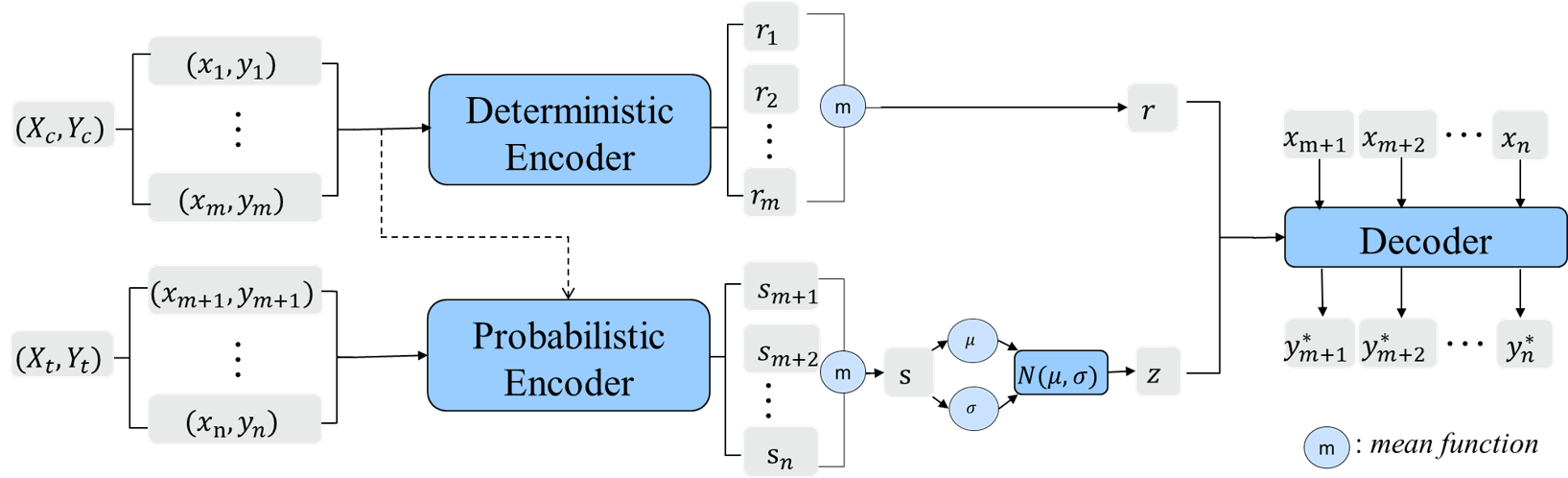}
    \caption{Flowchart of Neural Process.}
    \label{fig:1}
\end{figure*}

To model the distribution over random functions defined by (\ref{equ:6}), we build a NPBO that includes three components: a probabilistic encoder, a deterministic encoder, and a decoder. The mean function is applied after both probabilistic encoder and deterministic encoder by adding the vectors extracted from each encoder together and take the average. The overall network architecture in our implementation of NP is shown in Fig. \ref{fig:1}. As it can be seen, the probabilistic encoder is used to generate $\bm z$ that can be used to define the random process $f$. In this procedure, we use a NN to output $\mu$ and $\sigma$, which are used to build the multivariate normal distribution, then sample $\bm z$ from this normal distribution. The deterministic encoder with outcome $\bm r$ is to help improve the model stability. Finally, the decoder takes $x$, $\bm r$ and $\bm z$ as the inputs and predicts $\hat y=f(x)$. 

NP modify the evidence lower-bound (ELBO) \cite{yang2017understanding} to design the loss function. ELBO is given by:
\begin{equation}\label{equ:7}
    \scalebox{0.85}{
    $\begin{multlined}
        \text{log}p(y_{1:N}|x_{1:N}) \\
        \geq E_{q(\bm z|x_{1:N}, y_{1:N})}
            \left[\sum_{i=1}^{N}\text{log}p(y_i^*|f_{\bm z}(x_i))+
                \text{log}\frac{p(\bm z)}{q(\bm z|x_{1:N}, y_{1:N})} \right]
    \end{multlined}
    $
    }
\end{equation}
where $q(\bm z|x_{1:N}, y_{1:N})$ is a posterior of the latent vector $\bm z$.

In the training process, we split the data in each batch $\{x_1, ..., x_n\}$ into two sets: context points $(X_c, Y_c) = \{(x_1, y_1), ..., (x_m, y_m)\}$ and target points $(X_t, Y_t) = \{ (x_{m+1},y_{m+1}), ..., (x_n, y_n) \}$. The context points are fed to the deterministic encoder to produce the global representation of $\bm r$ and the target points are fed to the probabilistic encoder to generate $\bm s$ that generates $\bm z$. In this process, we assume that we only have the information of the context points, and want to predict  the target points. After this data splitting process, ELBO becomes:
\begin{equation}\label{equ:8}
    \scalebox{0.85}{$
    \begin{multlined}
        \text{log}p(Y_t | X_t, X_c, Y_c) \\
        \geq E_{q(\bm z|X_t, Y_t)}
            \left[\sum_{i=m+1}^{n}\text{log}p(y_i^*|f_{\bm z, \bm r}(x_i))+
                \text{log}\frac{p(\bm z|X_c, Y_c)}{q(\bm z|X_t, Y_t)} \right]
    \end{multlined}$
    }
\end{equation}

The final loss function is modified from the ELBO defined by (\ref{equ:8}). Because it is impossible to estimate the prior distribution of latent vector, i.e. $p(\bm z)$, so we use the posterior of context points $q(\bm z|X_c, Y_c)$ to approximate $p(\bm z|X_c, Y_c)$:
\begin{equation}\label{equ:9}
    \scalebox{0.85}{$
    \begin{multlined}
        \text{log}p(Y_t | X_t, X_c, Y_c) \\
        \geq E_{q(\bm z|X_t, Y_t)}
            \left[\sum_{i=m+1}^{n}\text{log}p(y_i^*|f_{\bm z, \bm r}(x_i))+
                \text{log}\frac{q(\bm z|X_c, Y_c)}{q(\bm z|X_t, Y_t)} \right]
    \end{multlined}$
    }
\end{equation}
We use the lower bound of (\ref{equ:9}) as the loss function. In order to obtain $q(\bm z|X_c, Y_c)$, the context points will be fed into probabilistic encoder for only the forward process.
Note that (\ref{equ:9}) contains two terms. The first is the expected log-likelihood over the target points. To calculate the first term, we need the context parameter $r$ and a sample from the latent space $z \sim q(z|X_t, Y_t)$, then feed $\bm z$, $\bm r$ and $X_t$ to the decoder to get the prediction of the model performance as well as its uncertainty. The second term evaluates the approximate negative Kullback–Leibler (KL) divergence \cite{Joyce2011} between $q(\bm z|X_c, Y_c)$ and $q(\bm z|X_t, Y_t)$, since we replace the prior of $\bm z \sim p(\bm z)$ to the posterior $\bm z \sim q(\bm z|X_c, Y_c)$.
In the inference process, we use all points observed as the context points, and $z$ is generated using this context points instead of target points; then perform the forward step of the training process to get the prediction which will be fed to Equ. (\ref{equ:ei}) to determine the next point to query.

Similar to GP, NP models distributions over functions and provides uncertainty estimations. Therefore, it is very suitable to be applied as the surrogate model under the Bayesian optimization framework. In other words, NP learns an implicit kernel from the data directly, which reduces the human effort to design the kernel function in GP, and leads to uncertainty estimations over unknown parameters. NP also combines benefits of neural networks so that is scaled linearly with respect to the number of observations.

\section{Experiments}
In this section, we compare the performance of NPBO on a power system parameter optimization problem with random search, GP-based Bayesian optimization, DNGO and BOHAMIANN. We further compare the NPBO with benchmark problems including GP-based Bayesian optimization, Random forest based Bayesian optimization, DNGO and BOHAMIANN on seven Bayesian optimization benchmark problems. 

\subsection{Parameter Optimization for Power System}
\begin{figure}
    \includegraphics[width=0.5\textwidth]{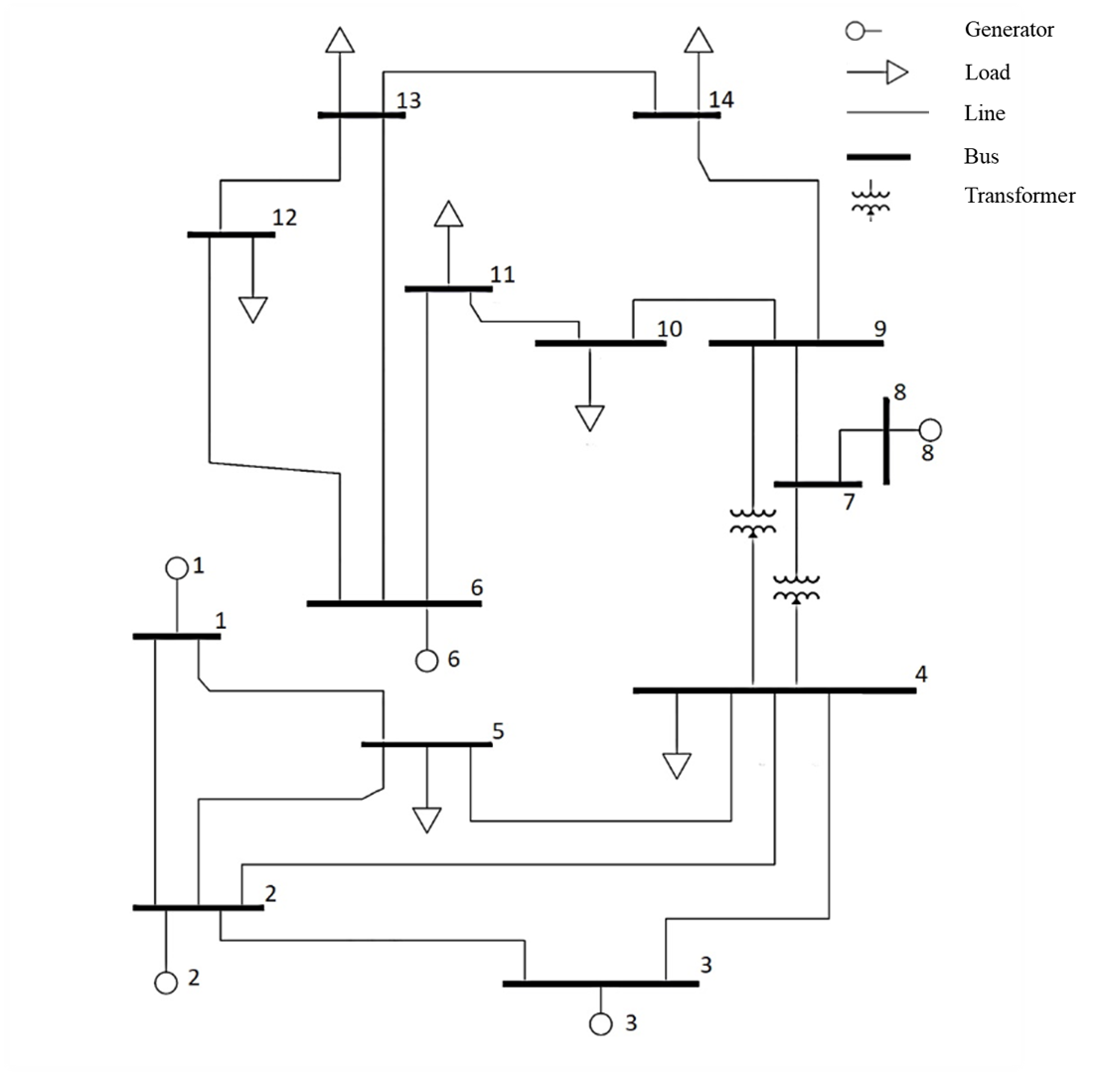}
    \caption{The IEEE 14 bus system.}
    \label{fig:14bus}
\end{figure}

Accurate and validated machine models are essential for reliable and economic power system operations. Machine models need to be regularly calibrated to ensure their accuracy for planning purpose and real-time operation \cite{huang2017calibrating}. In recent years, power generation is facing substantial changes to its power grid with increasing additions of renewable energy sources and generators. Consequently, it is critical to the system operators to have efficient calibration methods and tools in order to reduce the time and effort required in machine calibration.  
To test our proposed NPBO method, an IEEE 14-bus system \cite{IEEE14} with a 14-parameter generator model, namely the ROUND ROTOR GENERATOR MODEL (GENROU) shown in Fig.\ref{fig:14bus}, is simulated using the power system simulation tool, PSS\textcircled{\scriptsize{R}}E \cite{PSSE}. The simulator takes the 14-dimensional parameter of a generator model as the input, where their physical meanings are shown in Table \ref{tab:input}, to generate a 4-dimensional output measured on the bus terminal that is connected to the target generator, i.e., Bus Voltage Magnitude (voltage), Bus Voltage Frequency (frequency), Real Power Injection (P) and Reactive Power Injection (Q). 
To reduce the parameter dimension, the Design of Experiments (DOE), a well-established statistical approach, has been applied to select a subset of four out of 14 parameters \cite{gunawan2011fine}. That is, our goal is to optimize the selected four parameters so that the simulated  output  matches the target observations.  The ranges of input parameters to be optimized and observations are shown in Table \ref{tab:par}.

\begin{table}
    \caption{Input parameters and their physical meaning}
    \begin{center}
    \begin{tabular}{c|l}
    \hline
    $T'do$ & d-axis OC Transient time constnat\\
    $T''do$ & Subtransient\\
    $T'qo$ & Transient\\
    $T''qo$ & Subtransient\\
    $H$ & H Inertia \\
    $D$ & D speed Daping\\
    $Xd$ & Direct Axis Reactance\\
    $Xq$ & Quadrature Axis Reactance\\
    $X'd$ & Direct Axis Transient Reactance\\
    $X'q$ & Quadrature Axis Transient Reactance\\
    $X''d/X''q$ & Subtransient Reactance\\
    $XI$ & Leakage Reactance\\
    $S(1.0)$ & Saturation First Point\\
    $S(2.0)$ & Saturation Second Point\\
    \hline
    \end{tabular}
    \label{tab:input}
    \end{center}
\end{table}

\begin{table}
    \caption{Input and Output range of the power system}
	\begin{center}
		\begin{tabular}{|c|c|c|c|}
			\hline
			 \multicolumn{1}{|}{Parameter} & &  Min & Max \\
			\hline\hline
			\multirow{4}{*}{Input} & 
			$T'do$ & $5.625$ & $9.375$\\
		    & $Xd$  & $1.425$ & $2.375$\\
			& $Xq$ & $1.35$  & $2.25$ \\
			& $X'd$  & $0.315$  & $0.525$\\
			\hline
            \multirow{4}{*}{Output} & 
			$P$ & $-2.4612$ & $1.8012$\\
		    & $Q$  & $-0.5418$ & $8.3188$\\
			& $frequency$ & $-8.5615$  & $6.8365$ \\
			& $voltage$ & $0.9987$  & $1.0001$\\
			\hline
		\end{tabular}
	\end{center}
	
	\label{tab:par}
\end{table}

We use Mean Square Error (MSE) of the estimated parameters to evaluate the performance:
\begin{equation}\label{equ:mse}
    MSE = \frac{1}{Tr\times D}\sum_{i=1}^{Tr}\sum_{j=1}^{D} (P_{ij}-\hat{P_{ij}})^2
\end{equation}
where $Tr$ is the number of trials (e.g., each trial is initialized with a new set of ground-truth parameters), $D$ is the number of parameters to be optimized, i.e., the parameter dimension, $P_{ij}$ and $\hat P_{ij}$ represent the ground-truth and estimated $j^{th}$ parameter in $i^{th}$ trial, respectively. The term ground-truth denotes to the parameter combination that generates expected outputs. In our experiment, the objective function that needs to be minimized is defined by
\begin{equation}\label{equ:obj}
    D = \sum_{i=1}^m ||O_{i} - \hat{O}_{i} ||_2
\end{equation}
where $m=4$ is the dimension of the output, $O_{i}$ and $\hat{O}_{i}$ are vectors of length $452$ that represent outputs of ground-truth and of estimated parameters respectively.
We set $Tr=100$, and in each run, we query $500$ observations. The experiments are run on an Intel i7-9700k, and the results are shown in Table \ref{tab:2}. As residual block \cite{he2016deep} has seen great success in NNs, we re-implement a DNGO with two residual blocks and add it for comparison. As the table shows, NPBO has the most accurate parameter calibration  among all the models with a very short execution time.

\begin{figure}
    \includegraphics[width=0.5\textwidth, height=5cm]{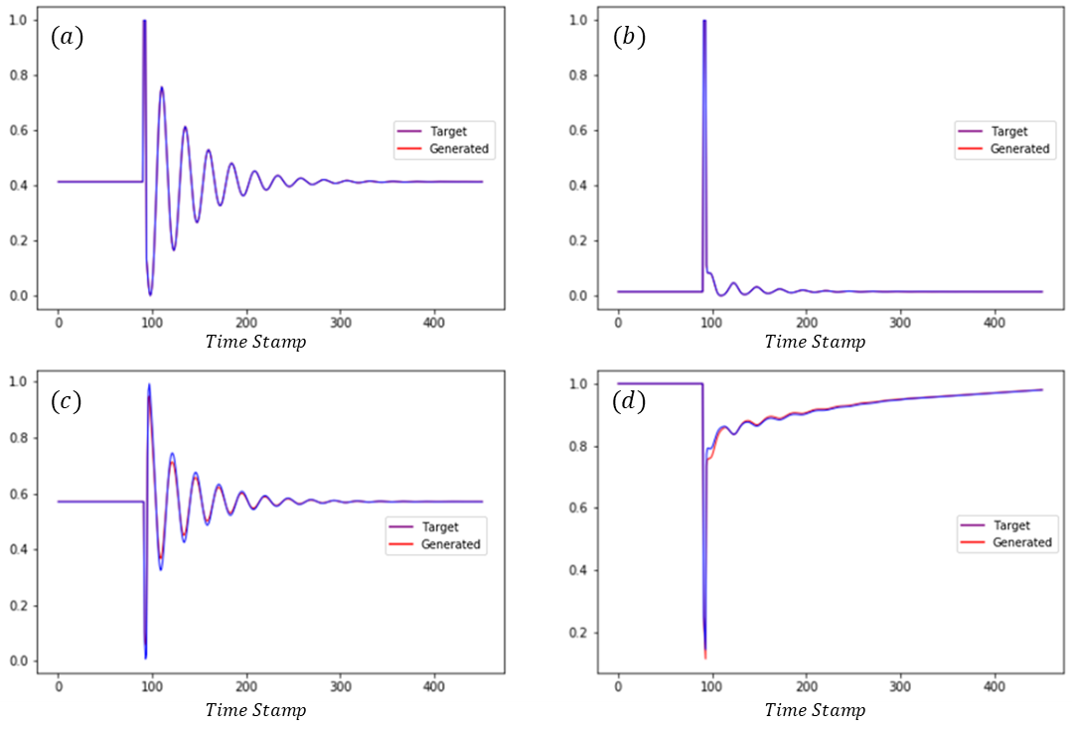}
    \caption{Comparison of Output from Calibrated Model and Target Output. (a): Real power injection; (b): Reactive power injection; (c): Frequency; (d) Voltage. Outputs are normalized using Min-Max Normalization method.}
    \label{fig:out}
\end{figure}

We further compare the four-dimensional outputs of the power system with the optimized parameters using NPBO to the observed target outputs in Fig. \ref{fig:out} . As it can be seen, there is only a very small difference between our optimized output and the target output, which indicates, with only a few observations, we can still obtain accurate and satisfactory parameter values.

\begin{table}
    \caption{Evaluation of Different Parameter Optimization Methods for power system}
    \begin{center}
    \begin{tabular}{c|c|c}
    \hline
    Experiment & Time(s) & MSE  \\
    \hline
    Random Search & $20$ & $2.001$ \\
    GP & $349$ & $1.759\times10^{-1}$ \\
    DNGO & $409$ & $1.504\times10^{-1}$ \\
    DNGO(Residual) & $997$ & $1.758\times10^{-1}$\\
    BOHAMIANN & $1672$ & $2.718\times10^{-2}$ \\
    NPBO & $157$ & $5.182\times10^{-3}$\\

    \hline
    \end{tabular}
    \label{tab:2}
    \end{center}
\end{table}

\subsection{Seven Benchmark Problems}
\begin{table*}[h]
    \caption{Evaluation of Different Surrogate Models on global optimization benchmarks}
    \begin{center}
    \setlength{\tabcolsep}{12pt}
    \resizebox{\textwidth}{50pt}{
    \begin{tabular}{c|c|c|c|c|c}
    \hline
    Experiment & Gaussian Process & Random Forest & DNGO & BOHAMIANN & NPBO \\
    \hline
    Branin & $\bm{0.3996}$ & $0.4562$ & 
    $0.4019$ & $\bm{0.3979}$ & $\bm{0.3980}$ \\
    
    Camelback & $\bm{-1.011}$ & $-0.8085$ & $\bm{-1.026}$ & $\bm{-1.027 }$ & $-0.9999 $\\
    
    Hartmann3 & $-1.028$ & $-0.998$
    & $\bm{-3.862}$ & $\bm{-3.861}$ & $\bm{-3.498}$ \\
    
    Forrester & $\bm{-6.021}$ & $\bm{-6.021}$
    & $-5.846$ & $\bm{-6.021}$ & $-5.301$ \\
    
    GoldsteinPrice & $\bm{4.916}$ & $27.69$ & $\bm{6.379}$ & $11.39$ & $\bm{8.654}$\\
    
    Hartmann6 & $\bm{-3.255}$ & $-3.132$
    & $\bm{-3.249}$ & $\bm{-3.264 }$ & $-3.214$ \\
    
    SinOne & $\bm{0.04292}$ & $0.06472$
    & $\bm{0.04292}$ & $0.04292 $ & $\bm{0.04292 }$ \\
    \hline
    \multicolumn{6}{l}{Top-3 best algorithms for each benchmark problem are bolded.}
    \end{tabular}}
    \label{tab:1}
    \end{center}
\end{table*}
To demonstrate the effectiveness of our approach, we compare NPBO to the state-of-the-art Bayesian optimization methods using different surrogate models on a set of synthetic functions \cite{klein2017robo}. Besides GP-based Bayesian optimization, DNGO and BOHAMIANN, we also use Random Forest \cite{hutter2011sequential} as the surrogate model in comparison. The goal is to find the parameters that minimize the synthetic functions. The results based on seven benchmark problems \cite{eggensperger2013towards} are shown in Table \ref{tab:1}. These benchmark problems are popular synthetic functions with the number of parameters range from one to six, e.g., Branin and Hartmann function \cite{surjanovic2013virtual}. As the table shows, all surrogate models with Bayesian optimization achieved acceptable performance. Overall, among the seven benchmark problems, NPBO performs competitive to BOHAMIANN, DNGO and GP based Bayesian optimization on four problems.

\begin{figure}
    \includegraphics[width=0.45\textwidth, height=5.5cm]{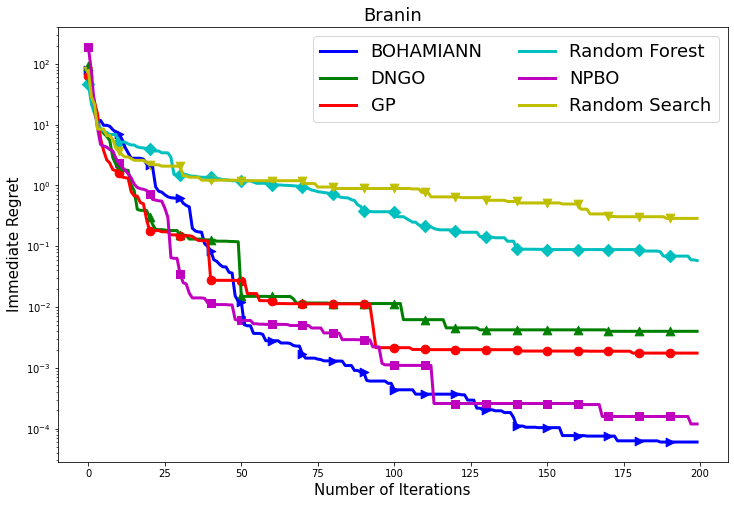}
    \caption{Immediate regret of different surrogate models applied to Bayesian Optimization on the Branin benchmark. Result averaged over 10 runs.}
    \label{fig:eva}
\end{figure}

We further show the optimization process of \textit{Branin} in detail in Fig. \ref{fig:eva}, where the performance is measured by immediate regret defined by (\ref{equ:ir})
\begin{equation}\label{equ:ir}
    I = |\hat f^i_{opt} - f_{opt}|
\end{equation}
where $\hat f^i_{opt}$ is the optimal observed value found in $i^{th}$ iteration, and $f_{opt}$ represents the theoretical optimal value. As it can be seen, NPBO performs competitively with BOHAMIANN and GP based Bayesian optimization, and its performance exceeds random search and Bayesian optimization based on random forest.

\section{Conclusions and Future Work}
In this paper, we propose Neural Process for Bayesian Optimization (NPBO) as a scalable parameter optimization method. NPBO has the ability to efficiently identify the optimal parameter combination of black-box models. The proposed model preserves the advantage of the GP such as flexibility and estimation of uncertainty, while reduces the time complexity from cubic to linear and improves the accuracy in uncertainty estimation.
NPBO is applied to optimize the parameters of a complex 14-parameter generator models in an IEEE 14-bus power system and the results show that NPBO outperforms the other benchmark algorithms, i.e., Gaussian Process, Random Forest, Deep Neural network for Global Optimization (DNGO) and Bayesian Optimization with Hamiltonian Monte Carlo Artificial Neural Networks (BOHAMIANN).
We further compared the performance of NPBO on seven common benchmark problems with different surrogate models and the results show NPBO has competitive performance with benchmark approaches. 

We consider three aspects in our future work: i) we are going to apply NPBO in different scenarios, e.g., accelerating experiments in the physical science \cite{bomlps2020}; ii) we will test the performance of variants of NP as the surrogate model, such as Conditional Neural Process \cite{garnelo2018conditional} and Attentive Neural Process \cite{kim2019attentive}; iii) acquisition function could also be replaced by NN to perform the trade-off strategy under Bayesian optimization framework.


\bibliography{reference}

\end{document}